\documentclass{article}

\usepackage{microtype}
\usepackage{graphicx}
\usepackage{subfigure}
\usepackage{booktabs} % for professional tables

\usepackage{hyperref}

\usepackage[accepted]{icml2023}

\usepackage{amsmath}
\usepackage{amssymb}
\usepackage{mathtools}
\usepackage{amsthm}

\usepackage[capitalize,noabbrev]{cleveref}

\theoremstyle{plain}

\theoremstyle{definition}

\theoremstyle{remark}

\usepackage[textsize=tiny]{todonotes}

\usepackage[T1]{fontenc}
\usepackage{kotex}

\usepackage[utf8]{inputenc}
\usepackage{todonotes}
\usepackage{microtype}
\usepackage{amssymb}
\usepackage{multirow}
\usepackage{multicol}
\usepackage{footnote}
\usepackage{tablefootnote}
\usepackage{inconsolata}
\usepackage{cuted}
\usepackage{caption}
\usepackage{graphicx}
\graphicspath{ {images/} }

\icmltitlerunning{Submission and Formatting Instructions for ICML 2023}

\begin{document}

\twocolumn[
% \icmltitle{A Call for  Error Explainable Benchmark (EEB) Dataset in Korean Automatic Speech Recognition Post-Processor}

\icmltitle{Toward Practical Automatic Speech Recognition and Post-Processing: \\ a Call for Explainable Error Benchmark Guideline}

% It is OKAY to include author information, even for blind
% submissions: the style file will automatically remove it for you
% unless you've provided the [accepted] option to the icml2023
% package.

% List of affiliations: The first argument should be a (short)
% identifier you will use later to specify author affiliations
% Academic affiliations should list Department, University, City, Region, Country
% Industry affiliations should list Company, City, Region, Country

% You can specify symbols, otherwise they are numbered in order.
% Ideally, you should not use this facility. Affiliations will be numbered
% in order of appearance and this is the preferred way.
\icmlsetsymbol{equal}{*}

\begin{icmlauthorlist}
\icmlauthor{Seonmin Koo}{yyy,equal}
\icmlauthor{Chanjun Park}{yyyc,equal}
\icmlauthor{Jinsung Kim}{yyy}
\icmlauthor{Jaehyung Seo}{yyy}
\icmlauthor{Sugyeong Eo}{yyy}
\icmlauthor{Hyeonseok Moon}{yyy}
\icmlauthor{Heuiseok Lim}{yyy}
%\icmlauthor{}{sch}
%\icmlauthor{}{sch}
\end{icmlauthorlist}

\icmlaffiliation{yyy}{Department of Computer Science and Engineering, Korea University, Seoul 02841, Korea}
\icmlaffiliation{yyyc}{Upstage, Gyeonggi-do, Korea}

\icmlcorrespondingauthor{Heuiseok Lim}{limhseok@korea.ac.kr}

% You may provide any keywords that you
% find helpful for describing your paper; these are used to populate
% the "keywords" metadata in the PDF but will not be shown in the document
\icmlkeywords{Machine Learning, ICML}

\vskip 0.3in
]

% this must go after the closing bracket ] following \twocolumn[ ...

% This command actually creates the footnote in the first column
% listing the affiliations and the copyright notice.
% The command takes one argument, which is text to display at the start of the footnote.
% The \icmlEqualContribution command is standard text for equal contribution.
% Remove it (just {}) if you do not need this facility.

%\printAffiliationsAndNotice{}  % leave blank if no need to mention equal contribution
\printAffiliationsAndNotice{\icmlEqualContribution} 
% otherwise use the standard text.

\begin{abstract}
Automatic speech recognition (ASR) outcomes serve as input for downstream tasks, substantially impacting the satisfaction level of end-users. Hence, the diagnosis and enhancement of the vulnerabilities present in the ASR model bear significant importance. However, traditional evaluation methodologies of ASR systems generate a singular, composite quantitative metric, which fails to provide comprehensive insight into specific vulnerabilities. This lack of detail extends to the post-processing stage, resulting in further obfuscation of potential weaknesses. Despite an ASR model's ability to recognize utterances accurately, subpar readability can negatively affect user satisfaction, giving rise to a trade-off between recognition accuracy and user-friendliness. To effectively address this, it is imperative to consider both the speech-level, crucial for recognition accuracy, and the text-level, critical for user-friendliness. Consequently, we propose the development of an Error Explainable Benchmark (EEB) dataset. This dataset, while considering both speech- and text-level, enables a granular understanding of the model's shortcomings. Our proposition provides a structured pathway for a more `real-world-centric' evaluation, a marked shift away from abstracted, traditional methods, allowing for the detection and rectification of nuanced system weaknesses, ultimately aiming for an improved user experience.

\end{abstract}

\section{Introduction}
Automatic speech recognition (ASR) is a task that recognizes voice and converts it into text, and it is getting more and more attention with the development of voice interface applications and devices such as Alexa, Siri, and Cortana~\citep{williams2007partially,wang2018modelling,wang2020data}. Although the performance of ASR models has improved with the development of technology, there are still things to be improved. Since speech recognition results are used as inputs for downstream tasks and affect end-user satisfaction, it is essential to accurately diagnose errors in the ASR model and improve them based on them~\citep{serdyuk2018towards,feng2022asr}. 

However, changing the structure of the ASR model itself or retraining it requires a considerable amount of data and is time-consuming and costly. Moreover, resources to train models are often scarce in the real world~\citep{park2020decoding}. Regarding this situation, post-processing research is emerging as a role to complement the ASR model. The ASR post-processing (ASRP) task aims to correct errors by detecting them in speech recognition results and can improve performance without changing the model structure~\citep{mani2020asr, liao2020improving,leng2021fastcorrect}. Therefore, effective diagnosis is required to enhance both ASR and ASRP models.

In the real world, there is a trade-off relationship between the ASR model's recognition accuracy and user-friendliness. Even if the ASR model accurately recognizes the input, the user's satisfaction may decrease. This is because humans do not always utter perfect sentences in the real world (e.g., incomplete utterances, sighs, etc.), so readability may deteriorate even if the ASR model accurately recognizes speech. The goal of the ASR model is to recognize the given voice input as text accurately, and the purpose of the ASRP model is to receive the recognition result as input and improve readability to increase end-user satisfaction. To achieve balanced speech recognition results in this trade-off situation, it is recommended to consider both the speech-level for recognition accuracy and the text-level for user-friendliness.
To this end, accurate diagnosis and evaluation of errors must be preceded in order to consider both speech- and text-level.

However, there are several challenges in the effective diagnosis and validation of the ASR and the ASRP models. First, the ASR model's existing evaluation methods are insufficient to diagnose the model's performance properly. Most ASR automatic evaluation processes are evaluated through quantitative metrics such as WER~\citep{woodard1982} and CER~\citep{morris2004}. However, since the metrics only deal with text, which is the result of voice recognition, there is a problem that the speech-level is not considered. Also, it is not easy to diagnose each model precisely because the existing benchmark datasets do not classify the characteristics of the collected voice data. In other words, each ASR model with different characteristics can receive the same quantitative score. However, the integrated evaluation score lacks the explanatory power to improve the model's weakness specifically. This leads to the use of human evaluation in the real world, even though many benchmark datasets have been released. Recently, attempts have been made to build data by considering the noisy environment or speaker characteristics to increase explanatory power~\citep{sikasote-anastasopoulos-2022-bembaspeech,lakomkin2019kt,gong2022vocalsound,dai2022ci}. However, these data also lack the explanatory ability to diagnose problems in the model because the number of error types used is limited.

In addition, the existing ASRP research also uses the ASR dataset, not a separate specialized dataset for learning, so those mentioned above low explanatory capability problem remains~\citep{sodhi2021mondegreen,kumar2022linguistically}. Although a grammatical error correction (GEC) dataset in which error types are subdivided exists, it is difficult to use it as it is in a speech recognition situation because it does not consider the speech recognition situation~\citep{koo2022k,yoon2022towards}. In summary, the diagnosis and verification of both ASR and ASRP are significant, but no proper segmentation benchmark exists.

Therefore, to alleviate this problem, we present an Error Explainable Benchmark (EEB) guideline for diagnosing and validating models by segmenting error types while considering both speech- and text-level. Such research is insufficient even in high-resource English and, in particular, does not exist in Korean. This is the first novel, a well-balanced guideline for both speech- and text-level. In addition, by building a dataset in consideration of the voice environment based on the existing Korean GEC dataset, it provides convenience in construction and enables real-world-centric evaluation.

\section{Related Work}
\subsection{Post-processing Model}
Post-processing serves an important role in quality enhancement across various fields by modifying the distorted output into appropriate statements. For instance, in the field of optical character recognition (OCR), traditional approaches such as manual, lexical, and statistical methods have been used~\citep{evershed2014correcting,nguyen2018adaptive}. More recently, language models like BERT have been employed for error detection in tasks like named entity recognition (NER) and are performed through character-level machine translation~\citep{nguyen2020neural}.

As another field, machine translation (MT) often utilizes the following methods. Post-processing research is being carried out in automatic post-editing (APE) to improve translation quality by adopting transfer learning~\citep{correia2019simple}. Concurrently, in the grammatical error correction (GEC) field, transformers and the copy mechanism are used to correct spelling and grammatical errors in MT results~\citep{lee2021korean}. Studies that define error types to construct test sets or utilize an automatic grammatical error annotation system to create datasets also exist to improve Korean GEC studies~\citep{koo2022k,yoon2022towards}. Likewise, the study on post-processing is actively explored in a wide range of fields and holds significance in terms of enhancing the quality of output results. This can also be of significant importance in the field of Automatic speech recognition (ASR), which is discussed in the following section.

\subsection{ASR Post-Processing Model}
ASR post-processing (ASRP) involves the detection and correction of errors in the output of an ASR, distinguishing it from simple error correction in that it considers user-friendliness as an additional aspect. This approach can improve the final quality of statements without modifying the ASR system structure. For instance, in specialized fields like the medical domain, attempts have been made to eliminate punctuation errors in ASR through post-processing~\citep{mani2020towards}. Prior research has primarily focused on providing information that allows humans to manually rectify erroneous segments, proposing alternative words for correction or creating an environment conducive to modification~\citep{suhm2001multimodal,feng2004using}. External information, such as word alternative hypothesis, noisy context, and accurate context, is provided to assist in post-processing for error correction~\citep{shi2011supporting}.
In particular, \citet{bassil2012asr} use the N-gram dataset for ASR errors to detect and correct errors automatically. Models such as LSTM-based or Transformer-based sequence-to-sequence architectures are adopted to correct the speech recognition results while considering the semantics and spelling~\citep{guo2019spelling,hrinchuk2020correction}.

Recent studies strive to improve ASRP performance by utilizing the results derived from ASR. \citet{gekhman-etal-2022-red} introduce the ASR confidence embedding (ACE) layer to the encoder of the ASR model to jointly encode the confidence scores and transcribed text into a contextualized representation. To mitigate the time and cost-related challenges associated with the parallel data required for training, \citet{park2021bts} employ Text-to-speech (TTS) and Speech-to-text (STT) technologies to construct parallel data.

\subsection{ASR dataset}
The availability of suitable datasets is imperative for the active progression of ASRP. Previously, post-processing studies have been conducted with ASR datasets. \citet{panayotov2015librispeech} organize the two labels in the ASR dataset that denote the quality of speech recognition, classified into `clean' and `other' categories, providing valuable assistance in the analysis. \citet{ardila-etal-2020-common} constuct comprehensive ASR dataset that includes demographic metadata such as age, sex, and accent to provide a wider representation. 

Transcription hypotheses obtained by decoding audio data using an ASR model are used to align hypothesis words with the reference (correct) transcription. The process of labeling errors and non-errors is facilitated by employing the minimum edit distance~\citep{gekhman2022red}. In the context of Chinese language datasets, a significant dataset is available for speech recognition systems, labeled with audio devices and recording environments~\citep{bu2017aishell}. \citet{gekhman2022red} build a dataset by aligning hypothesis words with the reference (correct) transcription through a transcription hypothesis obtained by decoding audio data with an ASR model and labeling errors and nonerrors using minimum edit distance. In the context of Chinese, a large-scale dataset is available for speech recognition systems labeled with audio device information and recording environments~\citep{bu2017aishell}.

To mitigate the problem of insufficient training data, methodologies that synthesize data via data augmentation methods have been proposed~\citep{liao2022automatic}. However, the overall quality of the data is more crucial than the size. Specifically, the detailed datasets that consider both speech- and text-level like the real world are absent. Consequently, we aim to construct the ASR Post-Processing dataset, which contemplates audio- and text-level for the first time.

\section{Proposed Error Explainable Benchmark
(EEB) Dataset }
\subsection{Why EEB?}
Most speech benchmark datasets typically consider error types from either speech-level for recognition accuracy or text-level for user-friendliness. However, speech recognition encompasses both speech and text in its processing pipeline, necessitating consideration of both aspects to mirror real-world situations truly. Our motivation for proposing a guideline for an EEB dataset, which contains both aspects, is detailed below.

Firstly, in real-world speech recognition, it is essential to consider the accuracy of model and end-user satisfaction simultaneously. To facilitate this, we propose to map the accuracy of the ASR model to `speech-level errors' and end-user satisfaction to `text-level errors' to mitigate this inherent trade-off.

From the perspective of the accuracy of the ASR model, it should output the recognition results `as heard,' regardless of the quality of the user-provided input. Conversely, from the standpoint of the end-user receiving the result, satisfaction increases when the output is presented in a refined state, despite any errors in the initial input. For instance, if a speaker stammers during their speech, the ASR model would likely deem its output more accurate if it recognizes and outputs all the words uttered. However, this would likely result in lower readability from the user's perspective.

A post-processor for speech recognition, validated by text-level error types, can be utilized to mitigate this. This post-processor would process speech recognition results to generate and provide user-friendly outcomes. Consequently, a benchmark dataset that can consider both aspects is required to handle the trade-off situation inherent in actual speech recognition scenarios.

Secondly, there are insufficient error types for a detailed diagnosis. Since benchmarks measure performance with quantitative metrics, it's crucial to fine-grain characteristics for a more detailed diagnosis. In industry contexts, communication between model and service teams is critical. When there's an issue with the model, clear criteria for the data flywheel significantly facilitate communication. That is, distinguishing the error type criteria for speech- and text-level aids in detailed diagnosis for model improvement. However, traditional benchmark datasets lack sufficient error types for detailed model analysis, leading to extensive usage of human evaluation in real-world settings. Humans can cope using commonsense, even if the criteria are unclear, but existing benchmarks with limited error types fall short. Hence, to solve the explainability issue, we must define the error type criteria that consider both the speech- and text-level and create benchmarks to achieve human-level explainability.

Thirdly, speech- and text-level errors coexist in real-world speech recognition systems. The ASR system needs to convert voice input into text output, a process often accompanied by simultaneous speech-level errors, like noisy environment issues, and text-level errors, such as spelling mistakes. However, the conventional focus has been chiefly on speech-level errors, making it challenging to handle a wide range of cases that could occur in reality. As a real-world setting requires consideration of both speech and text, we build a more practical benchmark considering both perspectives to enhance the explainability of speech recognition results.

To facilitate this, we define errors at speech- and text-level and propose a guideline for building speech recognition resources and benchmarks, as well as post-processor benchmarks that consider various environments.

\begin{table*}[]
\centering
\resizebox{\textwidth}{!}{
\begin{tabular}{|lll|l|}
\hline
\multicolumn{3}{|c|}{\textbf{Error Type}} &
  \multicolumn{1}{c|}{\textbf{Description}} \\ \hline\hline
\multicolumn{1}{|l|}{} &
  \multicolumn{1}{l|}{} &
  Washer/dryer machine &
   \\ \cline{3-3}
\multicolumn{1}{|l|}{} &
  \multicolumn{1}{l|}{\multirow{-2}{*}{Home appliances}} &
  Vacuum cleaner &
  \multirow{-2}{*}{Difficulty in recognition due to ambient electrical appliance noise.} \\ \cline{2-4} 
\multicolumn{1}{|l|}{} &
  \multicolumn{1}{l|}{} &
  Motorcycle &
   \\ \cline{3-3}
\multicolumn{1}{|l|}{} &
  \multicolumn{1}{l|}{} &
  Siren &
   \\ \cline{3-3}
\multicolumn{1}{|l|}{} &
  \multicolumn{1}{l|}{\multirow{-3}{*}{Individual transportation}} &
  Honk &
  \multirow{-3}{*}{Difficulty in recognition due to surrounding individual transportation noise.} \\ \cline{2-4} 
\multicolumn{1}{|l|}{} &
  \multicolumn{1}{l|}{} &
  Road side &
   \\ \cline{3-3}
\multicolumn{1}{|l|}{} &
  \multicolumn{1}{l|}{\multirow{-2}{*}{Street}} &
  Crowd &
  \multirow{-2}{*}{Difficulty in recognition due to the surrounding street noise.} \\ \cline{2-4} 
\multicolumn{1}{|l|}{} &
  \multicolumn{1}{l|}{} &
  Conversation &
   \\ \cline{3-3}
\multicolumn{1}{|l|}{} &
  \multicolumn{1}{l|}{\multirow{-2}{*}{Cafe/restaurant}} &
  Non-conversation &
  \multirow{-2}{*}{Challenges in perception due to the noise in cafes/restaurants.} \\ \cline{2-4} 
\multicolumn{1}{|l|}{} &
  \multicolumn{1}{l|}{} &
  Traditional market &
   \\ \cline{3-3}
\multicolumn{1}{|l|}{} &
  \multicolumn{1}{l|}{\multirow{-2}{*}{Market/shopping mall}} &
  Shopping mall &
  \multirow{-2}{*}{Difficulties in perception caused by the noise in markets/shopping malls.} \\ \cline{2-4} 
\multicolumn{1}{|l|}{} &
  \multicolumn{1}{l|}{} &
  Subway platform &
   \\ \cline{3-3}
\multicolumn{1}{|l|}{} &
  \multicolumn{1}{l|}{} &
  Inside the subway &
   \\ \cline{3-3}
\multicolumn{1}{|l|}{} &
  \multicolumn{1}{l|}{} &
  Inside the train (STR/KTX) &
   \\ \cline{3-3}
\multicolumn{1}{|l|}{} &
  \multicolumn{1}{l|}{\multirow{-4}{*}{Public transportation}} &
  Inside the bus &
  \multirow{-4}{*}{Difficulty in recognition due to surrounding public transportation noise.} \\ \cline{2-4} 
\multicolumn{1}{|l|}{} &
  \multicolumn{1}{l|}{} &
  Train terminal waiting room &
   \\ \cline{3-3}
\multicolumn{1}{|l|}{} &
  \multicolumn{1}{l|}{\multirow{-2}{*}{Terminal}} &
  Bus terminal waiting room &
  \multirow{-2}{*}{Challenges in perception due to the noise at terminals.} \\ \cline{2-4} 
\multicolumn{1}{|l|}{} &
  \multicolumn{1}{l|}{} &
  Outdoor construction site &
   \\ \cline{3-3}
\multicolumn{1}{|l|}{} &
  \multicolumn{1}{l|}{\multirow{-2}{*}{Construction site}} &
  Indoor construction site &
  \multirow{-2}{*}{Difficulties in perception caused by the noise at construction sites.} \\ \cline{2-4} 
\multicolumn{1}{|l|}{} &
  \multicolumn{1}{l|}{} &
  processing process &
   \\ \cline{3-3}
\multicolumn{1}{|l|}{} &
  \multicolumn{1}{l|}{\multirow{-2}{*}{Factory}} &
  Assembly process &
  \multirow{-2}{*}{Difficulties in perception caused by the noise in factories.} \\ \cline{2-4} 
\multicolumn{1}{|l|}{} &
  \multicolumn{1}{l|}{} &
  Sound of rain &
   \\ \cline{3-3}
\multicolumn{1}{|l|}{} &
  \multicolumn{1}{l|}{\multirow{-2}{*}{Nature ambient}} &
  Sound of the waves &
  \multirow{-2}{*}{Challenges in perception due to natural ambient noise.} \\ \cline{2-4} 
\multicolumn{1}{|l|}{\multirow{-24}{*}{Noisy environments}} &
  \multicolumn{1}{l|}{Etc.} &
  Artificial mechanical sound &
  \begin{tabular}[c]{@{}l@{}}In cases where external noise is present, although not falling into the \\ aforementioned categories.\end{tabular} \\ \hline
\multicolumn{1}{|l|}{} &
  \multicolumn{2}{l|}{Pause (silent)} &
  \begin{tabular}[c]{@{}l@{}}When there is a presence of pauses between syllables in speech that has \\ not yet concluded.\end{tabular} \\ \cline{2-4} 
\multicolumn{1}{|l|}{} &
  \multicolumn{2}{l|}{Filled pause} &
  When habitual sounds are inserted during moments of silence or break time. \\ \cline{2-4} 
\multicolumn{1}{|l|}{} &
  \multicolumn{2}{l|}{Interjection} &
  \begin{tabular}[c]{@{}l@{}}When phrases or longer segments are inserted regardless of their relevance \\ to the intended content being expressed.\end{tabular} \\ \cline{2-4} 
\multicolumn{1}{|l|}{} &
  \multicolumn{2}{l|}{Parenthetical} &
  \begin{tabular}[c]{@{}l@{}}When grammatically acceptable sentences are inserted without conveying \\ specific meaning or significance.\end{tabular} \\ \cline{2-4} 
\multicolumn{1}{|l|}{} &
  \multicolumn{2}{l|}{Unfinished interlocutor} &
  When speech is terminated without concluding the sentence. \\ \cline{2-4} 
\multicolumn{1}{|l|}{} &
  \multicolumn{2}{l|}{Word repetition} &
  Repeating the same word or phrase in succession during speech. \\ \cline{2-4} 
\multicolumn{1}{|l|}{} &
  \multicolumn{2}{l|}{Syllable repetition} &
  Repeating the same syllable in succession during speech. \\ \cline{2-4} 
\multicolumn{1}{|l|}{} &
  \multicolumn{2}{l|}{Phoneme repetition} &
  Repeating the same phoneme in succession during speech. \\ \cline{2-4} 
\multicolumn{1}{|l|}{} &
  \multicolumn{2}{l|}{Sustained} &
  When elongating certain parts of words within a sentence during speech. \\ \cline{2-4} 
\multicolumn{1}{|l|}{} &
  \multicolumn{2}{l|}{Hyperfluency} &
  When excessively verbose speech is employed. \\ \cline{2-4} 
\multicolumn{1}{|l|}{} &
  \multicolumn{2}{l|}{Mutter} &
  When muttering with an unclear demeanor. \\ \cline{2-4} 
\multicolumn{1}{|l|}{} &
  \multicolumn{2}{l|}{Dynamic error} &
  \begin{tabular}[c]{@{}l@{}}When syllabic intonation is inappropriate for the intended speech purpose \\ or difficult for human-level comprehension.\end{tabular} \\ \cline{2-4} 
\multicolumn{1}{|l|}{\multirow{-13}{*}{Characteristics of interlocutor}} &
  \multicolumn{2}{l|}{Speaking rate} &
  \begin{tabular}[c]{@{}l@{}}When speech rate is excessively fast, making it difficult for human-level \\ comprehension.\end{tabular} \\ \hline
\end{tabular}}
\caption{Proposed novel speech-level error type classification criteria for ASR and post-processing dataset}
\label{tab:speech_error_type}
\end{table*}

\begin{table*}[]
\centering
\resizebox{\textwidth}{!}{
\begin{tabular}{|lll|l|}
\hline
\multicolumn{3}{|c|}{\textbf{Error Type}} &
  \multicolumn{1}{c|}{\textbf{Description}} \\ \hline\hline
\multicolumn{3}{|l|}{Spacing} &
  Violating the spacing rules. \\ \hline
\multicolumn{3}{|l|}{Punctuation} &
  Punctuation marks are not attached in Korean sentences or are attached in the wrong. \\ \hline
\multicolumn{3}{|l|}{Numerical} &
  \begin{tabular}[c]{@{}l@{}}Cardinal number indicating quantity and the ordinal number indicating the order are \\ in error.\end{tabular} \\ \hline
\multicolumn{1}{|l|}{\multirow{11}{*}{Spelling and Grammatical}} &
  \multicolumn{2}{l|}{Remove} &
  Some words are not recognized, or endings or suffixes are omitted. \\ \cline{2-4} 
\multicolumn{1}{|l|}{} &
  \multicolumn{2}{l|}{Addition} &
  Same word is repeated, or an unused postposition or ending is added. \\ \cline{2-4} 
\multicolumn{1}{|l|}{} &
  \multicolumn{2}{l|}{Replace} &
  Word is replaced by another word. \\ \cline{2-4} 
\multicolumn{1}{|l|}{} &
  \multicolumn{2}{l|}{Separation} &
  Separating consonants and vowels in characters. \\ \cline{2-4} 
\multicolumn{1}{|l|}{} &
  \multicolumn{2}{l|}{\multirow{2}{*}{Foreign word conversion}} &
  Writing differently from the standard foreign language pronunciation. \\ \cline{4-4} 
\multicolumn{1}{|l|}{} &
  \multicolumn{2}{l|}{} &
  Instances of Incorrect Conversion of Syllables between English and Korean. \\ \cline{2-4} 
\multicolumn{1}{|l|}{} &
  \multicolumn{1}{l|}{\multirow{2}{*}{Spelling}} &
  Grapheme-to-phoneme(G2P) &
  Writing spellings according to pronunciation. \\ \cline{3-4} 
\multicolumn{1}{|l|}{} &
  \multicolumn{1}{l|}{} &
  Consonant vowel conversion &
  Spelling error in non-speaking alphabet units. \\ \cline{2-4} 
\multicolumn{1}{|l|}{} &
  \multicolumn{2}{l|}{Post-position} &
  Instances of inconsistent or missing post-position usage in target utterances. \\ \cline{2-4} 
\multicolumn{1}{|l|}{} &
  \multicolumn{2}{l|}{Syntax} &
  Cases of grammatically accurate yet interpretatively ambiguous meanings. \\ \cline{2-4} 
\multicolumn{1}{|l|}{} &
  \multicolumn{2}{l|}{Neologism} &
  \begin{tabular}[c]{@{}l@{}}Instances of discrepancy between target and its similarity in meaning, pronunciation, \\ and absence in Korean lexicon.\end{tabular} \\ \hline
\end{tabular}}
\caption{Proposed text-level error type classification criteria for ASR and post-processing dataset}
\label{tab:text_error_type}
\end{table*}

\subsection{Speech-Level Error Type}
Error types at the speech-level refer to factors that trigger inaccuracies in speech recognition situations. For example, identical utterances may be challenging to recognize due to background noise~\citep{sikasote-anastasopoulos-2022-bembaspeech}. Additionally, even in quiet environments, individuals do not consistently articulate perfect sentences and each speaker has unique characteristics that may negatively influence speech recognition~\citep{gong2022vocalsound}.

Table~\ref{tab:speech_error_type} illustrates the speech-level error type classification criteria considering these characteristics. The speech-level error types allow the classification of two main categories (noisy environments and characteristics of interlocutor) and more detailed error types, with 24 sub-types for noise error and 13 for speaker characteristics.

Considering environments inundated with noise, it does not represent a quiet recording situation but rather a condition intertwined with noise. Real-world scenarios frequently involve inputs replete with ambient noise~\citep{sikasote-anastasopoulos-2022-bembaspeech}. Reflecting on these practical situations where voice interface applications and devices are deployed, we propose an enhanced categorization scheme that closely follows the classification in the AI-HUB's noisy environment speech recognition dataset \footnote{\url{https://www.aihub.or.kr/}} which are representative Korean data platform. We divide the noisy environment errors into 11 nuanced subcategories, including \textbf{home appliances}, where recognition is impaired due to surrounding appliance noise; \textbf{individual transportation}, which includes instances with ambient transportation noise; \textbf{street}, covering situations with disruptive street noise; \textbf{cafe/restaurant}, addressing cases with the cafe or restaurant ambient noise; \textbf{market/shopping mall}, indicating instances with market or shopping mall noise; \textbf{public transportation}, comprising cases with subway or bus noise; \textbf{terminal}, reflecting instances with terminal noise; \textbf{construction site}, for cases hindered by construction site noise; \textbf{factory}, indicating instances with factory noise; \textbf{nature ambient}, for cases disturbed by natural sounds. Lastly, we include an \textbf{etc.} category for instances where recognition is affected by external noise types not encompassed in the previous categories.

Considering speaker characteristics, recognition can be hampered due to the individual traits of the recorder. Inspired by studies on idiolectal elements in the field of psycholinguistics~\citep{ha2008,shin2005study}, we propose a nuanced categorization comprising 13 detailed subcategories. \textbf{Pause (silent)} category captures instances where silence intervenes mid-utterance before completion—for instance, when `I am eating' is articulated as `I am... eating'. \textbf{Filled pause} represents cases characterized by the habitual insertion of filler sounds during pauses, as in utterances supplemented by sounds such as `um... uh... so I'. \textbf{Interjection} category encompasses instances where one or more words or phrases irrelevant to the intended message are interjected, evident in utterances like `Okay I see, but you know'. \textbf{Parenthetical} category includes instances where grammatically correct, but semantically neutral phrases are inserted—for instance, utterances incorporating phrases such as `you know' and `I mean'. \textbf{Unfinished interlocutor} category denotes cases where the utterance concludes prematurely—for instance, when `I am eating' is truncated to `I am...'. \textbf{Word repetition} category signifies instances where the same word is iterated, as in saying `Hello' as `Hello Hello'. \textbf{Syllable repetition} category characterizes cases where the same syllable is iterated—for instance, when `Hello' is articulated as `He-hello'. \textbf{Phoneme repetition} category encapsulates instances where the same phoneme is repeated, such as saying `Hello' as `Hel-llo'. \textbf{Sustained} category accounts for instances where part of an utterance is elongated, exemplified in `Is that so—right?'. \textbf{Hyperfluency} category represents instances of excessive verbosity. \textbf{Mutter} category includes cases where utterances are murmured in an indistinct manner, as in `That.. is.. like that...'. \textbf{Dynamic error} category encompasses instances where syllable articulation strength is incongruous with the intended utterance, or instances that are challenging to comprehend at the human-level. Finally, \textbf{speaking rate} category accounts for instances where rapid speech pace hinders comprehension at a human-level.

\subsection{Text-Level Error Type}

Text-level error types refer to issues that emerge in speech recognition results and must be addressed by post-processing. Since the output of the speech recognizer serves as the input for downstream tasks, it is one of the most significant factors influencing end-user satisfaction. Enhancing end-user satisfaction is possible by improving input quality for downstream tasks and diagnosing post-processing model performance through detailed error types.

Existing datasets that detail error types, such as GEC datasets, do not consider speech recognition situations~\citep{koo2022k,yoon2022towards}. Therefore, we reconfigure the Korean GEC dataset, K-NCT, to suit speech recognition situations. The existing K-NCT dataset includes errors that only occur at the text-level and not in speech situations~\citep{koo2022k}. Hence, errors that do not have vocal characteristics are removed.

Table~\ref{tab:text_error_type} illustrates the text-level error type classification criteria considering speech recognition situations, including 13 text-level errors that can occur in speech recognition situations.

\textbf{Spacing} encapsulates instances contravening standard spacing conventions. \textbf{Punctuation} entails cases where punctuation is omitted or misapplied in Korean sentences—for instance, when `Can I teach?' is interpreted as `Can I teach.' \textbf{Numerical} encompasses cases where number conversion fails, such as when `Ahead of the three-month schedule' is interpreted as `Bill 2, 3-month schedule'.

\textbf{Spelling and Grammar} consists of ten detailed subcategories. \textbf{Remove} designates cases where some word components are not recognized, or endings or particles are missing—for example, when `The champion is in the final' is misinterpreted as `Champion final'. \textbf{Addition} involves cases where the same word is repeated or unutilized particles or endings are appended. For instance, when `World's fruits, fish, and meat' is interpreted as `World's world's fruits, fish, and meat'. \textbf{Replace} refers to instances where one word is substituted with another—for example, when `Apply the filter.' is interpreted as `Wear the pizza'. \textbf{Separation} refers to instances where consonants and vowels in the target utterance are separated, exemplified when `The discount applies as it is.' is interpreted as `Discount app - lise as it is.'. \textbf{Foreign word conversion} refers to cases where words deviate from standard foreign word pronunciation or some syllables are incorrectly converted from English to Korean or vice versa. For example, when `Brazil's Samba Festival' is interpreted as `Brazil's SsamBap Festival,' or `I prefer to use ATM.' is interpreted as `I prefer to use hm.'.

\textbf{Spelling} is bifurcated into two types: Grapheme-to-Phoneme (G2P) and Consonant vowel conversion. \textbf{G2P} pertains to instances where a character is recognized per its pronunciation. \textbf{Consonant vowel conversion} refers to instances where phonemic units are incorrectly spelled. \textbf{Post-position} refers to cases where different particles are used or omitted—for example, when `Ordinary high school students' is interpreted as `Ordinary at high school students.' \textbf{Syntax} involves cases where the grammatical interpretation remains valid, but the semantic interpretation varies. Finally, \textbf{neologism} refers to cases where the target word and its meaning and pronunciation are dissimilar and are not included in Korean vocabulary.

\section{Construction Process Design}

In this work, we propose a comprehensive data construction guideline for the ASR and ASRP dataset, grounded in the application of a grammatical error correction (GEC) dataset. Our methodology encompasses validation, speech recording, noise synthesis, and difficulty tagging of a GEC dataset featuring text-level discrepancies. For the efficiency of the task, we choose the `consensus labeling' method~\citep{tang2011semi}, in which a human overseer, who possesses an elevated degree of task completion, serves as a quality controller. During the progression of the task, any outcomes that do not conform to the established guidelines are promptly dismissed and subsequently reconstructed.

\subsection{Step 1: Verification of Text}
In this study, we employ a human-curated GEC dataset, which encompasses various text-level error types~\citep{koo2022k}. Considering the inapplicability of the standard GEC benchmark dataset in a speech recognition setting, we selectively compose text-level error types dataset. In particular, we extract 13 categories that resonate with speech recognition scenarios (e.g., honorific colloquial expression) and reorganize their hierarchy for ease of labeling. Consequently, our refined dataset includes data reflecting 13 error types relevant to speech recognition contexts.

Subsequently, we authenticate the quality of the filtered dataset focusing on the alignment between labels and text, and the inclusion of text-level errors with a specific consideration of the speech recognition context. Validation processes proceed with a human supervisor, priorily trained with each error type. Evaluators are presented with an erroneous sentence, its correct counterpart, and a specified error type with the corresponding error span indicated. They are then tasked with assessing whether the sentence contains the presented error types. Sentences deemed to be incorrect are appropriately amended. This procedural framework ensures the generation of a high-quality dataset.

\subsection{Step 2: Speech Recording}
In the second phase, we request the recording participants to incorporate characteristics of interlocutor errors into their recordings by presenting them with speech-level errors and transcription relevant to the respective error types. At most 3 error types are presented, which could include an instance of `no error type', indicating clean data. The placement of the error within the sentence is non-specific, with the ensurance that it includes only the errors specified. The recording environment should be ensured to be quiet without background noise. Each recorder is instructed to speak as naturally as possible, emulating their speech patterns when interacting with a voice interface application in real-world scenarios. After completing the recording, participants have the opportunity to listen to their own voice, and if they determine that the speech does not meet the criteria, they can re-record it. Participants are required to go through the process of listening to their recorded speech in order to complete the recording task.

\subsection{Step 3: Synthesis of Background Noise}
In the next stage, we incorporate background noise into the recording to reflect the noise environment error in the proposed speech-level. The background noise used for this integration is derived directly from recordings of the identified environments. We ensure that the collected noise spans a duration longer than that of the recording file, fostering noise diversity. To mimic real-world situations, we conduct both single and multiple noise syntheses while filtering out instances that are unlikely to co-occur. During noise synthesis, the noise is integrated as though it is ambient background noise, designed to be audible at the onset of the voice file. Noise is composited into the recording by randomly excising sections, thus ensuring variation within sounds, even when they are categorized under the same noise type.

\subsection{Step 4: Difficulty Annotation}
We employ a framework that distinguishes between utterances considered easier for ASR and those deemed harder or more noisy for ASR~\citep{breiner2022userlibri}. We extend this framework to include the tagging of difficulty using a Likert scale by human annotators. Humans listen to audio file and select score based on evaluation criteria. We ask humans, `How difficult is it to recognize the presented speech accurately as the same as the transcript?' Scores range from 1 (very easy) to 5 (very difficult). Three evaluators assess each audio file, and the average score is selected as the difficulty level of the data. This allows for a detailed analysis of the model's performance

\section{Conclusion}
Since speech recognition results are used as inputs for downstream tasks and affect end-user satisfaction, it is important to diagnose and improve the weak types of speech recognition models. Considering the trade-off in real-world scenarios, achieving a balanced ASR environment requires diagnosing and validating both speech-level accuracy and text-level user-friendliness. We propose an Error Explainable Benchmark (EEB) guideline for diagnosing and validating models by segmenting error types while considering both speech- and text-level. To facilitate the construction process, we utilize a GEC dataset that includes text-level errors and structure the process into validation, recording, synthesis of background noise, and difficulty tagging stages, employing consensus labeling within each stage to enhance the efficiency and quality of the task. Through our EEB guidelines, it is possible to build a dataset that can specifically diagnose and verify the performance of ASR models and post-processors. Through this, automatic evaluation close to the real-world situation is possible.

\section*{Acknowledgements}
This work was supported by Institute of Information \& communications Technology Planning \& Evaluation(IITP) grant funded by the Korea government(MSIT) (No. 2020-0-00368, A Neural-Symbolic Model for Knowledge Acquisition and Inference Techniques), and this work was supported by the National Research Foundation of Korea(NRF) grant funded by the Korea government(MSIT)(No. 2022R1A5A7026673).

% Acknowledgements should only appear in the accepted version.
% \section*{Acknowledgements}

% \textbf{Do not} include acknowledgements in the initial version of
% the paper submitted for blind review.

% If a paper is accepted, the final camera-ready version can (and
% probably should) include acknowledgements. In this case, please
% place such acknowledgements in an unnumbered section at the
% end of the paper. Typically, this will include thanks to reviewers
% who gave useful comments, to colleagues who contributed to the ideas,
% and to funding agencies and corporate sponsors that provided financial
% support.

% In the unusual situation where you want a paper to appear in the
% references without citing it in the main text, use \nocite
\nocite{langley00}

% \bibliography{example_paper}
\bibliography{custom}

\begin{thebibliography}{42}
\providecommand{\natexlab}[1]{#1}
\providecommand{\url}[1]{\texttt{#1}}
\expandafter\ifx\csname urlstyle\endcsname\relax
  \providecommand{\doi}[1]{doi: #1}\else
  \providecommand{\doi}{doi: \begingroup \urlstyle{rm}\Url}\fi

\bibitem[Ardila et~al.(2020)Ardila, Branson, Davis, Kohler, Meyer, Henretty,
  Morais, Saunders, Tyers, and Weber]{ardila-etal-2020-common}
Ardila, R., Branson, M., Davis, K., Kohler, M., Meyer, J., Henretty, M.,
  Morais, R., Saunders, L., Tyers, F., and Weber, G.
\newblock Common voice: A massively-multilingual speech corpus.
\newblock In \emph{Proceedings of the Twelfth Language Resources and Evaluation
  Conference}, pp.\  4218--4222, Marseille, France, May 2020. European Language
  Resources Association.
\newblock ISBN 979-10-95546-34-4.
\newblock URL \url{https://aclanthology.org/2020.lrec-1.520}.

\bibitem[Bassil \& Semaan(2012)Bassil and Semaan]{bassil2012asr}
Bassil, Y. and Semaan, P.
\newblock Asr context-sensitive error correction based on microsoft n-gram
  dataset.
\newblock \emph{arXiv preprint arXiv:1203.5262}, 2012.

\bibitem[Breiner et~al.(2022)Breiner, Ramaswamy, Variani, Garg, Mathews, Sim,
  Gupta, Chen, and McConnaughey]{breiner2022userlibri}
Breiner, T., Ramaswamy, S., Variani, E., Garg, S., Mathews, R., Sim, K.~C.,
  Gupta, K., Chen, M., and McConnaughey, L.
\newblock Userlibri: A dataset for asr personalization using only text.
\newblock \emph{arXiv preprint arXiv:2207.00706}, 2022.

\bibitem[Bu et~al.(2017)Bu, Du, Na, Wu, and Zheng]{bu2017aishell}
Bu, H., Du, J., Na, X., Wu, B., and Zheng, H.
\newblock Aishell-1: An open-source mandarin speech corpus and a speech
  recognition baseline.
\newblock In \emph{2017 20th conference of the oriental chapter of the
  international coordinating committee on speech databases and speech I/O
  systems and assessment (O-COCOSDA)}, pp.\  1--5. IEEE, 2017.

\bibitem[Correia \& Martins(2019)Correia and Martins]{correia2019simple}
Correia, G.~M. and Martins, A.~F.
\newblock A simple and effective approach to automatic post-editing with
  transfer learning.
\newblock \emph{arXiv preprint arXiv:1906.06253}, 2019.

\bibitem[Dai et~al.(2022)Dai, Cahyawijaya, Yu, Barezi, Xu, Yiu, Frieske,
  Lovenia, Winata, Chen, et~al.]{dai2022ci}
Dai, W., Cahyawijaya, S., Yu, T., Barezi, E.~J., Xu, P., Yiu, C.~T., Frieske,
  R., Lovenia, H., Winata, G., Chen, Q., et~al.
\newblock Ci-avsr: A cantonese audio-visual speech datasetfor in-car command
  recognition.
\newblock In \emph{Proceedings of the Thirteenth Language Resources and
  Evaluation Conference}, pp.\  6786--6793, 2022.

\bibitem[Evershed \& Fitch(2014)Evershed and Fitch]{evershed2014correcting}
Evershed, J. and Fitch, K.
\newblock Correcting noisy ocr: Context beats confusion.
\newblock In \emph{Proceedings of the First International Conference on Digital
  Access to Textual Cultural Heritage}, pp.\  45--51, 2014.

\bibitem[Feng \& Sears(2004)Feng and Sears]{feng2004using}
Feng, J. and Sears, A.
\newblock Using confidence scores to improve hands-free speech based navigation
  in continuous dictation systems.
\newblock \emph{ACM Transactions on Computer-Human Interaction (TOCHI)},
  11\penalty0 (4):\penalty0 329--356, 2004.

\bibitem[Feng et~al.(2022)Feng, Yu, Cai, Liu, Zheng, and Wang]{feng2022asr}
Feng, L., Yu, J., Cai, D., Liu, S., Zheng, H.-T., and Wang, Y.
\newblock Asr-robust spoken language understanding on asr-glue dataset.
\newblock 2022.

\bibitem[Gekhman et~al.(2022{\natexlab{a}})Gekhman, Zverinski, Mallinson, and
  Beryozkin]{gekhman-etal-2022-red}
Gekhman, Z., Zverinski, D., Mallinson, J., and Beryozkin, G.
\newblock {RED}-{ACE}: Robust error detection for {ASR} using confidence
  embeddings.
\newblock In \emph{Proceedings of the 2022 Conference on Empirical Methods in
  Natural Language Processing}, pp.\  2800--2808, Abu Dhabi, United Arab
  Emirates, December 2022{\natexlab{a}}. Association for Computational
  Linguistics.
\newblock URL \url{https://aclanthology.org/2022.emnlp-main.180}.

\bibitem[Gekhman et~al.(2022{\natexlab{b}})Gekhman, Zverinski, Mallinson, and
  Beryozkin]{gekhman2022red}
Gekhman, Z., Zverinski, D., Mallinson, J., and Beryozkin, G.
\newblock Red-ace: Robust error detection for asr using confidence embeddings.
\newblock \emph{arXiv preprint arXiv:2203.07172}, 2022{\natexlab{b}}.

\bibitem[Gong et~al.(2022)Gong, Yu, and Glass]{gong2022vocalsound}
Gong, Y., Yu, J., and Glass, J.
\newblock Vocalsound: A dataset for improving human vocal sounds recognition.
\newblock In \emph{ICASSP 2022-2022 IEEE International Conference on Acoustics,
  Speech and Signal Processing (ICASSP)}, pp.\  151--155. IEEE, 2022.

\bibitem[Guo et~al.(2019)Guo, Sainath, and Weiss]{guo2019spelling}
Guo, J., Sainath, T.~N., and Weiss, R.~J.
\newblock A spelling correction model for end-to-end speech recognition.
\newblock In \emph{ICASSP 2019-2019 IEEE International Conference on Acoustics,
  Speech and Signal Processing (ICASSP)}, pp.\  5651--5655. IEEE, 2019.

\bibitem[Ha \& Sim(2008)Ha and Sim]{ha2008}
Ha, J.-W. and Sim, H.~S.
\newblock A comparison study of interjectional characteristics between people
  who stutter and people who do not stutter.
\newblock \emph{Communication Sciences and Disorders}, 13\penalty0
  (3):\penalty0 438--453, 2008.

\bibitem[Hrinchuk et~al.(2020)Hrinchuk, Popova, and
  Ginsburg]{hrinchuk2020correction}
Hrinchuk, O., Popova, M., and Ginsburg, B.
\newblock Correction of automatic speech recognition with transformer
  sequence-to-sequence model.
\newblock In \emph{Icassp 2020-2020 ieee international conference on acoustics,
  speech and signal processing (icassp)}, pp.\  7074--7078. IEEE, 2020.

\bibitem[Koo et~al.(2022)Koo, Park, Seo, Lee, Moon, Lee, and Lim]{koo2022k}
Koo, S., Park, C., Seo, J., Lee, S., Moon, H., Lee, J., and Lim, H.
\newblock K-nct: Korean neural grammatical error correction gold-standard test
  set using novel error type classification criteria.
\newblock \emph{IEEE Access}, 10:\penalty0 118167--118175, 2022.

\bibitem[Kumar et~al.(2022)Kumar, Adiga, Ranjan, Krishna, Ramakrishnan, Goyal,
  and Jyothi]{kumar2022linguistically}
Kumar, R., Adiga, D., Ranjan, R., Krishna, A., Ramakrishnan, G., Goyal, P., and
  Jyothi, P.
\newblock Linguistically informed post-processing for asr error correction in
  sanskrit.
\newblock \emph{Proc. Interspeech 2022}, pp.\  2293--2297, 2022.

\bibitem[Lakomkin et~al.(2019)Lakomkin, Magg, Weber, and
  Wermter]{lakomkin2019kt}
Lakomkin, E., Magg, S., Weber, C., and Wermter, S.
\newblock Kt-speech-crawler: Automatic dataset construction for speech
  recognition from youtube videos.
\newblock \emph{arXiv preprint arXiv:1903.00216}, 2019.

\bibitem[Lee et~al.(2021)Lee, Shin, Lee, and Choi]{lee2021korean}
Lee, M., Shin, H., Lee, D., and Choi, S.-P.
\newblock Korean grammatical error correction based on transformer with copying
  mechanisms and grammatical noise implantation methods.
\newblock \emph{Sensors}, 21\penalty0 (8):\penalty0 2658, 2021.

\bibitem[Leng et~al.(2021)Leng, Tan, Zhu, Xu, Luo, Liu, Qin, Li, Lin, and
  Liu]{leng2021fastcorrect}
Leng, Y., Tan, X., Zhu, L., Xu, J., Luo, R., Liu, L., Qin, T., Li, X., Lin, E.,
  and Liu, T.-Y.
\newblock Fastcorrect: Fast error correction with edit alignment for automatic
  speech recognition.
\newblock \emph{Advances in Neural Information Processing Systems},
  34:\penalty0 21708--21719, 2021.

\bibitem[Liao et~al.(2020)Liao, Eskimez, Lu, Shi, Gong, Shou, Qu, and
  Zeng]{liao2020improving}
Liao, J., Eskimez, S.~E., Lu, L., Shi, Y., Gong, M., Shou, L., Qu, H., and
  Zeng, M.
\newblock Improving readability for automatic speech recognition transcription.
\newblock \emph{Transactions on Asian and Low-Resource Language Information
  Processing}, 2020.

\bibitem[Liao et~al.(2022)Liao, Shi, and Xu]{liao2022automatic}
Liao, J., Shi, Y., and Xu, Y.
\newblock Automatic speech recognition post-processing for readability: Task,
  dataset and a two-stage pre-trained approach.
\newblock \emph{IEEE Access}, 10:\penalty0 117053--117066, 2022.

\bibitem[Mani et~al.(2020{\natexlab{a}})Mani, Palaskar, and
  Konam]{mani2020towards}
Mani, A., Palaskar, S., and Konam, S.
\newblock Towards understanding asr error correction for medical conversations.
\newblock In \emph{Proceedings of the first workshop on natural language
  processing for medical conversations}, pp.\  7--11, 2020{\natexlab{a}}.

\bibitem[Mani et~al.(2020{\natexlab{b}})Mani, Palaskar, Meripo, Konam, and
  Metze]{mani2020asr}
Mani, A., Palaskar, S., Meripo, N.~V., Konam, S., and Metze, F.
\newblock Asr error correction and domain adaptation using machine translation.
\newblock In \emph{ICASSP 2020-2020 IEEE International Conference on Acoustics,
  Speech and Signal Processing (ICASSP)}, pp.\  6344--6348. IEEE,
  2020{\natexlab{b}}.

\bibitem[Morris et~al.(2004)Morris, Maier, and Green]{morris2004}
Morris, A., Maier, V., and Green, P.
\newblock From wer and ril to mer and wil: improved evaluation measures for
  connected speech recognition.
\newblock 01 2004.

\bibitem[Nguyen et~al.(2018)Nguyen, Coustaty, Doucet, Jatowt, and
  Nguyen]{nguyen2018adaptive}
Nguyen, T.-T.-H., Coustaty, M., Doucet, A., Jatowt, A., and Nguyen, N.-V.
\newblock Adaptive edit-distance and regression approach for post-ocr text
  correction.
\newblock In \emph{Maturity and Innovation in Digital Libraries: 20th
  International Conference on Asia-Pacific Digital Libraries, ICADL 2018,
  Hamilton, New Zealand, November 19-22, 2018, Proceedings 20}, pp.\  278--289.
  Springer, 2018.

\bibitem[Nguyen et~al.(2020)Nguyen, Jatowt, Nguyen, Coustaty, and
  Doucet]{nguyen2020neural}
Nguyen, T. T.~H., Jatowt, A., Nguyen, N.-V., Coustaty, M., and Doucet, A.
\newblock Neural machine translation with bert for post-ocr error detection and
  correction.
\newblock In \emph{Proceedings of the ACM/IEEE joint conference on digital
  libraries in 2020}, pp.\  333--336, 2020.

\bibitem[Panayotov et~al.(2015)Panayotov, Chen, Povey, and
  Khudanpur]{panayotov2015librispeech}
Panayotov, V., Chen, G., Povey, D., and Khudanpur, S.
\newblock Librispeech: an asr corpus based on public domain audio books.
\newblock In \emph{2015 IEEE international conference on acoustics, speech and
  signal processing (ICASSP)}, pp.\  5206--5210. IEEE, 2015.

\bibitem[Park et~al.(2020)Park, Yang, Park, and Lim]{park2020decoding}
Park, C., Yang, Y., Park, K., and Lim, H.
\newblock Decoding strategies for improving low-resource machine translation.
\newblock \emph{Electronics}, 9\penalty0 (10):\penalty0 1562, 2020.

\bibitem[Park et~al.(2021)Park, Seo, Lee, Lee, Moon, Eo, and Lim]{park2021bts}
Park, C., Seo, J., Lee, S., Lee, C., Moon, H., Eo, S., and Lim, H.-S.
\newblock Bts: Back transcription for speech-to-text post-processor using
  text-to-speech-to-text.
\newblock In \emph{Proceedings of the 8th Workshop on Asian Translation
  (WAT2021)}, pp.\  106--116, 2021.

\bibitem[Serdyuk et~al.(2018)Serdyuk, Wang, Fuegen, Kumar, Liu, and
  Bengio]{serdyuk2018towards}
Serdyuk, D., Wang, Y., Fuegen, C., Kumar, A., Liu, B., and Bengio, Y.
\newblock Towards end-to-end spoken language understanding.
\newblock In \emph{2018 IEEE International Conference on Acoustics, Speech and
  Signal Processing (ICASSP)}, pp.\  5754--5758. IEEE, 2018.

\bibitem[Shi \& Zhou(2011)Shi and Zhou]{shi2011supporting}
Shi, Y. and Zhou, L.
\newblock Supporting dictation speech recognition error correction: the impact
  of external information.
\newblock \emph{Behaviour \& Information Technology}, 30\penalty0 (6):\penalty0
  761--774, 2011.

\bibitem[Shin et~al.(2005)Shin, Ahn, Nam, and Kwon]{shin2005study}
Shin, M.-S., Ahn, J.-B., Nam, H.-W., and Kwon, D.-H.
\newblock A study of dysfluency characteristics in normal adults and children
  in monologue.
\newblock \emph{Speech Sciences}, 12\penalty0 (3):\penalty0 49--57, 2005.

\bibitem[Sikasote \& Anastasopoulos(2022)Sikasote and
  Anastasopoulos]{sikasote-anastasopoulos-2022-bembaspeech}
Sikasote, C. and Anastasopoulos, A.
\newblock {B}emba{S}peech: A speech recognition corpus for the {B}emba
  language.
\newblock In \emph{Proceedings of the Thirteenth Language Resources and
  Evaluation Conference}, pp.\  7277--7283, Marseille, France, June 2022.
  European Language Resources Association.
\newblock URL \url{https://aclanthology.org/2022.lrec-1.790}.

\bibitem[Sodhi et~al.(2021)Sodhi, Chio, Jash, Onta{\~n}{\'o}n, Apte, Kumar,
  Jeje, Kuzmin, Fung, Cheng, et~al.]{sodhi2021mondegreen}
Sodhi, S.~S., Chio, E. K.-I., Jash, A., Onta{\~n}{\'o}n, S., Apte, A., Kumar,
  A., Jeje, A., Kuzmin, D., Fung, H., Cheng, H.-T., et~al.
\newblock Mondegreen: A post-processing solution to speech recognition error
  correction for voice search queries.
\newblock In \emph{Proceedings of the 27th ACM SIGKDD Conference on Knowledge
  Discovery \& Data Mining}, pp.\  3569--3575, 2021.

\bibitem[Suhm et~al.(2001)Suhm, Myers, and Waibel]{suhm2001multimodal}
Suhm, B., Myers, B., and Waibel, A.
\newblock Multimodal error correction for speech user interfaces.
\newblock \emph{ACM transactions on computer-human interaction (TOCHI)},
  8\penalty0 (1):\penalty0 60--98, 2001.

\bibitem[Tang \& Lease(2011)Tang and Lease]{tang2011semi}
Tang, W. and Lease, M.
\newblock Semi-supervised consensus labeling for crowdsourcing.
\newblock In \emph{SIGIR 2011 workshop on crowdsourcing for information
  retrieval (CIR)}, pp.\  1--6, 2011.

\bibitem[Wang et~al.(2020)Wang, Fazel-Zarandi, Tiwari, Matsoukas, and
  Polymenakos]{wang2020data}
Wang, L., Fazel-Zarandi, M., Tiwari, A., Matsoukas, S., and Polymenakos, L.
\newblock Data augmentation for training dialog models robust to speech
  recognition errors.
\newblock \emph{arXiv preprint arXiv:2006.05635}, 2020.

\bibitem[Wang et~al.(2018)Wang, Gunter, and VanDyke]{wang2018modelling}
Wang, S., Gunter, T., and VanDyke, D.
\newblock On modelling uncertainty in neural language generation for policy
  optimisation in voice-triggered dialog assistants.
\newblock In \emph{2nd Workshop on Conversational AI: Today’s Practice and
  Tomorrow’s Potential, NeurIPS}, 2018.

\bibitem[Williams \& Young(2007)Williams and Young]{williams2007partially}
Williams, J.~D. and Young, S.
\newblock Partially observable markov decision processes for spoken dialog
  systems.
\newblock \emph{Computer Speech \& Language}, 21\penalty0 (2):\penalty0
  393--422, 2007.

\bibitem[Woodard \& Nelson(1982)Woodard and Nelson]{woodard1982}
Woodard, J. and Nelson, J.
\newblock An information theoretic measure of speech recognition performance.
\newblock 1982.

\bibitem[Yoon et~al.(2022)Yoon, Park, Kim, Cho, Park, Kim, Seo, and
  Oh]{yoon2022towards}
Yoon, S., Park, S., Kim, G., Cho, J., Park, K., Kim, G.~T., Seo, M., and Oh, A.
\newblock Towards standardizing korean grammatical error correction: Datasets
  and annotation.
\newblock \emph{arXiv preprint arXiv:2210.14389}, 2022.

\end{thebibliography}
\bibliographystyle{icml2023}

%%%%%%%%%%%%%%%%%%%%%%%%%%%%%%%%%%%%%%%%%%%%%%%%%%%%%%%%%%%%%%%%%%%%%%%%%%%%%%%
%%%%%%%%%%%%%%%%%%%%%%%%%%%%%%%%%%%%%%%%%%%%%%%%%%%%%%%%%%%%%%%%%%%%%%%%%%%%%%%
% APPENDIX
%%%%%%%%%%%%%%%%%%%%%%%%%%%%%%%%%%%%%%%%%%%%%%%%%%%%%%%%%%%%%%%%%%%%%%%%%%%%%%%
%%%%%%%%%%%%%%%%%%%%%%%%%%%%%%%%%%%%%%%%%%%%%%%%%%%%%%%%%%%%%%%%%%%%%%%%%%%%%%%
\newpage
\appendix
\onecolumn
% \section{You \emph{can} have an appendix here.}

% You can have as much text here as you want. The main body must be at most $8$ pages long.
% For the final version, one more page can be added.
% If you want, you can use an appendix like this one, even using the one-column format.
%%%%%%%%%%%%%%%%%%%%%%%%%%%%%%%%%%%%%%%%%%%%%%%%%%%%%%%%%%%%%%%%%%%%%%%%%%%%%%%
%%%%%%%%%%%%%%%%%%%%%%%%%%%%%%%%%%%%%%%%%%%%%%%%%%%%%%%%%%%%%%%%%%%%%%%%%%%%%%%

\end{document}